# Beyond “AI Language”: The case for the idiolectal nature of LLM output

Karolina Rudnicka
Faculty of Languages,
University of Gdańsk,
Poland
karolina.rudnicka@ug.edu.pl

Thomas Stephan Juzek
Department of Scientific Computing,
Florida State University,
United States
tjuzek@fsu.edu

**Abstract**

While large language model outputs are frequently analysed as a collective super variety termed “AI language,” this chapter argues that this perspective coexists with distinct, model-specific linguistic signatures akin to human idiolects. We analyse two datasets of LLM-generated texts on societal topics: a 2024 corpus of six models (Improta et al. 2024) and a newly generated 2026 corpus using the same prompts featuring six contemporary models. Our findings, utilising computational descriptors and stylometric principal component analysis reveal a generational shift between the style of the 2024 and 2026 cohorts, while demonstrating that each individual model maintains a unique linguistic profile. This multi-layered interplay is illustrated by contraction frequencies, which vary from over 1,200 to over 30,000 per million words within the same cohort of models (2026). Ultimately, we conclude that treating LLM output as idiolectal in nature provides a valuable framework with potential implications for research on variation and change, LLM-generated text detection, forensic linguistics and usage-based approaches to language.

## 1. Introduction

Since the emergence of widely available, consumer-oriented tools powered by artificial intelligence (AI), a growing number of scholars have focused on comparing the linguistic output of large language models (LLMs) – most notably ChatGPT-3.5 and ChatGPT-4 – with human-authored texts (i.a., Reviriego et al. 2024a; Georgiou 2025; Zanotto & Aroyehun 2024; Zhang & Crosthwaite 2026). These studies broadly aim to investigate the capabilities of LLMs, their lexical properties, and the characteristics of AI-generated text. Given that such tools are trained on large-scale, heterogeneous corpora, the underlying assumption in much of this research appears to be that AI-generated text reflects human writing as a collective phenomenon. Accordingly, AI-generated texts are often compared with texts produced by numerous authors. Systematic documentation of their linguistic backgrounds and proficiency levels is, nonetheless, frequently absent from such comparisons, possibly because of data constraints. This assumption is, however, rarely made explicit. In Reviriego et al. (2024b), for instance, a subset of 126 TOEFL essays is used as a basis for the human-authored comparison; yet no detailed sociolinguistic information of the essay authors is provided, leaving the makeup of the human baseline unclear[1].

A similar lack of specificity is noted in Georgiou (2025), where the human-authored comparison consists of five IELTS writing task samples produced by professional teachers. This design choice does helpfully narrow the otherwise heterogeneous category of human-authored text by anchoring it to a clearly defined task and writer group – yet further information about the teachers' linguistic backgrounds remains unavailable.

Taken together, such approaches implicitly treat the textual output of different LLMs as a single super-variety, *AI language*, and human-authored texts, regardless of authors' background, as another, *human language*. From the perspective of linguistic variation studies, this risks oversimplification. For instance, in established subfields of English linguistics such as World Englishes, dialect studies, second language acquisition, English as a lingua franca, and diachronic linguistics, a more granular approach to language variety is standard practice. The field of AI language investigation, still in its formative stages, has yet to adopt a comparably differentiated analytical lens.

At the same time, research has shown that individual AI-powered tools have stable and recurring linguistic characteristics (e.g., Kobak et al. 2025; Stokel-Walker 2024). These include, among others, a preference for standard grammatical forms and a systematic dispreference for slang or non-standard linguistic features (Smith et al. 2025; Rodriguez Louro 2025; Amin et al. 2025; Mair 2026), and less lexical diversity and the presence of more nominalisations when compared to human-written texts (André et al. 2023; Herbold et al. 2023; Reviriego et al. 2024; Muñoz-Ortiz et al. 2024, Terčon & Dobrovoljc 2025). Also, according to Zamaraeva et al. (2025: 9047), "a human-authored text is more similar to an LLM-generated text than to another human-authored text" which means that human-human variation is greater than human-LLM variation. Nevertheless, as will be argued in the following section, the prevailing analytical framing of this very new and rapidly growing research field raises questions that have so far not received sufficient attention – questions about what precisely is being compared to what, and whether the categories employed are analytically adequate.

[1] Also, no information can be extracted from the files in the repository from which the essays were sourced (Shijaku 2022).

The present chapter examines the linguistic nature of the textual output generated by LLM-based AI tools and argues that the output from a particular AI tool represents a language variety similar to what linguists refer to as an *idiolect* defined, e.g., as "the habits of expression of a single person" (Encyclopaedia Britannica; more definitions in the next section). The chapter is structured as follows. Section 2 provides an in-depth discussion of the distinction between the mutually non-exclusive notions of *AI language* and *AI idiolects*, framing them as complementary analytical perspectives with distinct theoretical implications. It sets out both the rationale for and the limitations of applying the notion of idiolect – or AI-idiolect[2] – to the output of LLM-based tools. Section 3 presents our approach and an overview of research questions that our empirical analysis addresses, while Sections 4–6 support our argument through tailored analyses drawing on computational, stylometric, and corpus-linguistic methods. Section 7 contains the conclusions.

## 2. From AI language to AI idiolects

A search for 'AI language' on ResearchGate returns over 3,500 pages of research output – preprints, published works, datasets, and more – reflecting the scale of scholarly activity at the intersection of artificial intelligence and linguistics. Many of the works encountered when searching databases such as ResearchGate or Google Scholar address the application of AI-powered tools in teaching, learning, and student assessment (e.g., Dai & Liu 2024; Kim 2023; Owan et al. 2023; Singh & Hiran 2022), or the opportunities and risks AI poses for education

[2] Also referred to as *chatolect* in the previous preprint of one of the Author's (Rudnicka 2025b).

(e.g., Hsiao et al. 2025). However, as noted in the Introduction, there is also a growing body of research on the linguistic characteristics of what may be termed a super-variety: AI language – referred to variously in the literature as LLM-generated, AI-written, AI-generated, or machine-generated text, among other labels.

This terminological variation is itself symptomatic of a deeper conceptual underspecification. As Geng & Poibeau (2025: 2) demonstrate, existing definitions of LLM-generated text vary considerably in scope and precision, ranging from broad formulations such as that of Crothers et al. (2023: 70979) – "machine-generated text is natural language text that is produced, modified, or extended by a machine" – to more technically precise ones, such as Kumarage et al. (2024: 2): "we define AI-generated text as output produced by a natural language generation pipeline employing a neural probabilistic language model", and Wu et al. (2025: 278): "[LLM-generated text] is defined as cohesive, grammatically sound, and pertinent content generated by LLMs". For the purposes of clarity and analytical manageability, the present chapter adopts the definition proposed by Wu et al. (2025: 278).

Terčon & Dobrovoljc (2025) provide a systematic survey of 44 studies examining the linguistic characteristics of AI-generated text (AIGT) in comparison to human-written text (HWT). Investigating texts across lexical, grammatical, and other levels of linguistic description, they synthesise findings on methodology, models, genres, languages, and prompting approaches. Among their most consistently reported results is that AIGT tends toward a more formal and impersonal style – evidenced by higher frequencies of nouns, determiners, and adpositions – alongside lower lexical diversity and a tendency toward repetitive text. They also note that existing research remains heavily concentrated on English and on the GPT model family, and that prompt sensitivity is rarely addressed: of the 44 studies surveyed, only 15 (approximately 34%) draw on data from more than one model family. This observation sets the scene for the argument developed in the following subsection.

### 2.1 Idiolect as a fitting notion

The overrepresentation of ChatGPT in existing research – noted by Terčon & Dobrovoljc (2025) and reflected in further works examined in the present chapter (i.a., Reviriego et al. 2024; Georgiou 2025; Zanotto & Aroyehun 2024; Zhang & Crosthwaite 2026) – raises a concern: if the majority of studies treat AI language as a category while drawing almost exclusively on ChatGPT data, what they are effectively describing is the language of a single model, not a super-variety. This focus is not without justification: during the period in which most of these studies were conducted, ChatGPT held an overwhelming share of the AI assistant market, making it a reasonable proxy for AI-generated text (SensorTower 2026). Nevertheless, for a super-variety to be analytically meaningful, it would need to be grounded in data from several model families, examined collectively against a comparable human baseline.

This observation may shift analytical attention from the collective category to the individual tool. At the lexical level, the LexA-Index AI Word Overuse Explorer (Juzek 2026) documents overused and spiking words per individual AI-powered tool – and these lists differ across tools. This phenomenon finds a natural parallel in the linguistic field of idiolectal lexicography. As Wojan (2024: 22) notes, the term author's lexicography is ambiguous: it may be understood either as lexicography (1) whose object of interest is the language of a specific author, or (2) that has its own author, i.e. one created or practised by a given individual. With regard to the object of study, Wojan (2024: 23) proposes the term *idiolectal lexicography* as a notion relative to *idiolectal dictionary* – that is, a dictionary of an individual's language or of the idiolect of a particular author. The systematic variation across models in lexical overuse documented by Juzek (2026) suggests that something analogous may be warranted for individual LLM-based tools.

In a previous exploratory study (Rudnicka 2025), corpus-linguistic and stylometric methods – namely Burrows's Delta method – were applied to test whether individual LLM-based tools exhibit stable and distinguishable linguistic profiles. Using a dataset of texts on diabetes generated by ChatGPT and Gemini (Naveed 2024), she found that a random ten percent sample of ChatGPT texts yields a stylometric distance of 0.92 to the full ChatGPT dataset and 1.49 to the Gemini dataset, while a corresponding Gemini sample shows a distance of 0.84 to Gemini and 1.45 to ChatGPT – a pattern consistent with two distinct authorial voices rather than one. That study frames this finding explicitly in terms of idiolect, proposing that individual LLM-based tools express themselves in ways analogous to the individual speaking styles that linguists traditionally attribute to human speakers. Additionally, the two models appear to have different preferences concerning choice of vocabulary – with ChatGPT preferring academic and formal ways of communicating, and Gemini using more informal and everyday expressions (as illustrated by the different frequencies of nouns such as *sugar* preferred by Gemini, and *glucose* chosen much more frequently by ChatGPT).

This line of research has recently converged toward what Milička et al. (2025: 2) describe as an emerging consensus: "ChatGPT and Gemini default personas exhibit their own unique writing styles, akin to human idiolects" (a conclusion they support by results from Bitton et al. (2025), Rao et al. (2025), and Rudnicka (2025a,b)).

The notion of idiolect is well established in the linguistic literature. Litvinova (2021: 14) defines it as "an individual way of language use which is formed under the influence of a wide range of factors". A more comprehensive definition is offered by Richards, Platt & Platt (1992: 172):

> Idiolect – the language system of an individual as expressed by the way he or she speaks or writes within the overall system of a particular language. In its widest sense, someone's idiolect includes their way of communicating; for example, their choice of utterances and the way they interpret the utterances made by others. In a narrower sense, an idiolect may include those features, either in speech or writing, which distinguish one individual from others (…).

Crucially, an idiolect is understood as emergent rather than chosen: it arises from an individual's history of language use and stays relatively stable across topics and contexts, while remaining unique to that individual. This distinguishes it from the concept of style, which denotes deliberate choices about formality or tone – choices that can be adopted, adjusted, or imitated. As Richards, Platt & Platt (1992: 360) discusses it in the definition of *stylistics*:

> the study of that variation in language (STYLE) which is dependent on the situation in which the language is used and also on the effect the writer or speaker wishes to create on the reader or hearer. (…) Stylistics is concerned with the choices that are available to a writer and the reasons why particular forms and expressions are used rather than others.

If the linguistic characteristics of LLM-based tools are consistent, stable, model-specific, and present on many levels (i.a. lexical, syntactic, semantic), they fit more naturally to the concept of idiolect than to style.

Adopting an idiolect-oriented perspective toward the textual output of LLM-powered tools leads to meaningfully different conclusions than treating all AI-generated text as a single super-variety. For instance, lexical diversity and richness are properties of every language user's idiolect, shaped among other things by educational background and language exposure. It is an important question whether the lexical diversity of ChatGPT-4 reflects the average language use of a population, or whether it represents an individual competence unique to that model version – one shaped by, among other factors, the type, quality, timing of its training data. Preliminary findings by Reviriego et al. (2024a) for ChatGPT-3.5 and ChatGPT-4 appear to support the latter interpretation.

This perspective also opens onto broader theoretical questions relevant to construction grammar (i.a., Dunn 2024; Hilpert 2024; Bergs 2025; Fried & Nikiforidou 2025), usage-based approaches to language (i.a., Diessel 2017; Pemberton 2024; Dunn 2024), language change (i.a., Kytö, Scahill, Tanabe 2010; Hickey 2012; Rudnicka 2023; Baker 2023; Millar 2024), and discourse-pragmatic processes such as colloquialisation (i.a., Mair 1998; Rühlemann & Hilpert 2017). The fundamental question underlying all of these is whether, after training, an LLM-based tool merely averages and mirrors its training data, or whether individual lexical, grammatical, and syntactic habits crystallise in the process – habits stable enough to constitute something that can be meaningfully described as an idiolect.

## 3. Research questions, data, and analytical approach

The present chapter is built around three overarching research questions:
RQ1: In what sense is the notion of idiolect applicable to the output of LLM-based AI tools?
RQ2: On which linguistic levels do individual LLM-based tools exhibit stable and consistent differences in their output?
RQ3: Do AI-idiolects change over time – and can we identify family-level continuities between successive model generations that occupy a middle ground between the super-variety AI language and the idiolect of a particular tool?

To address these research questions, the chapter draws on two datasets. The first is TextualLLMap (Improta et al., 2024), a corpus of 28,000 texts produced by GPT-3.5, GPT-4o, Haiku, Llama-3-8B, Llama-3.1-70B, and Mistral-7b on societal topics, representing LLM-generated language as of 2024. Five of the six models were released in 2024, with Mistral being the exception, having been released in late 2023. The second is a newly generated dataset produced in 2026 using exactly the same prompts and topics; with generation parameters as close as possible to those used by Improta et al. (2024). It features data obtained from three open-weight models (Mistral-Nemo 12B, OLMo-3 7B, Qwen-3 14B) and three closed-API models (Haiku-4.5, Gemini-3 Flash, GPT-5.4 Mini).

Due to the use of these two cohorts (one containing data generated in 2024 and one containing data from 2026), a direct cross-temporal comparison is possible for some of the models, in particular the different versions of GPT (3.5, 4.0 in the 2024 cohort, and 5.4 Mini from the 2026 cohort) and Claude Haiku (3.5 from 2024 and 4.5 from late 2025, belonging to the 2026 cohort). To ensure comparability, the Mistral data-generation procedure from Improta et al. (2024) is replicated using the legacy model in the current version of LM Studio. The datasets will be made publicly available upon publication of the present chapter.

The analytical focus is deliberately broad rather than focused on lexis. While vocabulary constitutes one dimension of investigation, the chapter prioritises more general linguistic descriptors – including output length, sentence length, and the frequency of disclaimers (see Section 4) – alongside stylometric analysis via principal component analysis (PCA), see Section 5, and corpus-linguistic methods focused on contractions, presented in Section 6. This combination allows us to assess whether, and to what degree, individual LLM-based tools exhibit stable, distinguishable linguistic profiles – that is, whether the notion of AI- or LLM-idiolect is empirically warranted.

To the best of our knowledge, the present study is also the first to adopt an explicitly diachronic perspective, comparing the linguistic characteristics of different model generations both individually and as temporally defined groups.

### 3.1 Generation of the 2026 corpus: replication and extension

Our general approach is to have different models produce text from the same prompts, and then run various linguistic analyses. Different models producing text from the same prompts has been done before; for example, Improta et al. (2024) analyse model generations at the level of topic analysis. Their generations are essays about “climate”, “global warming”, “maths anxiety”, and “misinformation around health”.

From the six models in Improta et al. (2024) – GPT-3.5, GPT-4o, Claude-3.5-Haiku, Llama-3-8B, Llama-3.1-70B, and Mistral-7B – we take their generations for our linguistic analyses[3]. Because their generations are mostly from 2024 models (with one exception of Mistral-7B being a 2023 model), we refer to them as the 2024 cohort. In order to offer a look at more recent data, we also add output from more recent models and refer to them as the 2026 cohort. Further, for validity, we first rerun one of their runs to validate our generations, and use this as a bridge: we rerun Mistral-7B via LM Studio 0.4.11 on an RTX 4070 Ti Super, so that we have Mistral-7B generations from the 2024 Improta et al. (2024) data, and fresh 2026 generations. We then add further current models with 2026 data: Gemini-3 Flash, GPT-5.4 Mini, and Claude-4.5-Haiku are closed models that we queried via API; OLMo-3-7B, Mistral-Nemo-12B, and Qwen-3-14B are open models that we ran with vLLM on an A100 80 GB using Lambda. Most models are from late 2025 and early 2026, with the exception of Mistral-Nemo-12B, which is from mid-2024 and was included as the then current successor to Mistral-7B. We keep the four topics and, just as in Improta et al. (2024), generate 1,000 texts per topic.

[3] The datasets generated by Improta et al (2024) also contain datasets in Italian. In the present work, we only use their datasets in English.

We took the topics, the exact prompts, and, where known, the settings from Improta et al. (2024): temperature 0.5 and top-p 0.95. Top-k, repetition penalty, frequency penalty, and presence penalty use default parameters; if there is any variation, it is to be expected here. The sampling cap is 2,048 tokens, with $n \leq 1{,}000$ per topic; there are rare occasions of model refusal. The system prompt is: "Below is an instruction that describes a task. Write a response that appropriately completes the request." This identical string is used across all topics and models. Closed-API runs use the same string in the 'system' role; for the open-weights '--no-system-role path', we concatenate it onto the user turn. For reasoning models, and this only concerns Qwen3-14B in our case, we disabled reasoning.

Especially older models sometimes keep generating. Since we set an upper limit for answers, we therefore use a truncation detector: word count $\geq 0.80 \times$ cap and no terminal punctuation. Real cap hits concentrate in GPT-5.4 Mini (4-16%, depending on topic); all other models are below 0.5%. The cap is 2,048 tokens. We apply a regex/heuristic pre-parser for cleaning and exclude from the length and lexical metrics the following: repetition loops (1.6% of texts), code contamination (1.3%), and non-English output (~0%); about 3% of the generations in total. We keep the following, which we report below in more detail: disclaimer openers, AI self-reference, hedging openers, and refusals. The exact script can be found in our code repository.

## 4. Computational methods – the broad picture

The present section presents a broad computational description of the two model cohorts. Using an NLP pipeline, we examine four descriptors, namely i) output and sentence length, ii) type-token ratio and vocabulary size, iii) presence of disclaimers and AI self-reference behaviour, and iv) a brief comparison with human reference data.

### 4.1 Methodology: pipeline and descriptors

We process all data through the same pipeline: basic cleaning (filtering of duplicates, code, repetition loops, and non-English replies), part-of-speech tagging with spaCy (v3.8.7; Honnibal et al., 2020), and dependency parsing with 'en_core_web_lg'.

We then provide initial descriptives, where we analyse model families and individual models using the 2024 vs 2026 split from above. We analyse for (i) answer length and average sentence length, (ii) type-token ratio and corpus vocabulary size, (iii) disclaimer behaviour, and (iv) a brief comparison with related human behaviour.

### 4.2 Results

With respect to cleaning, the vast majority of model generations are fine. However, Mistral shows heightened repetition in health-misinformation responses in both the 2024 and 2026 data, at around 12-14%. See Figure 4.1 for details. We observe notable variation across models.

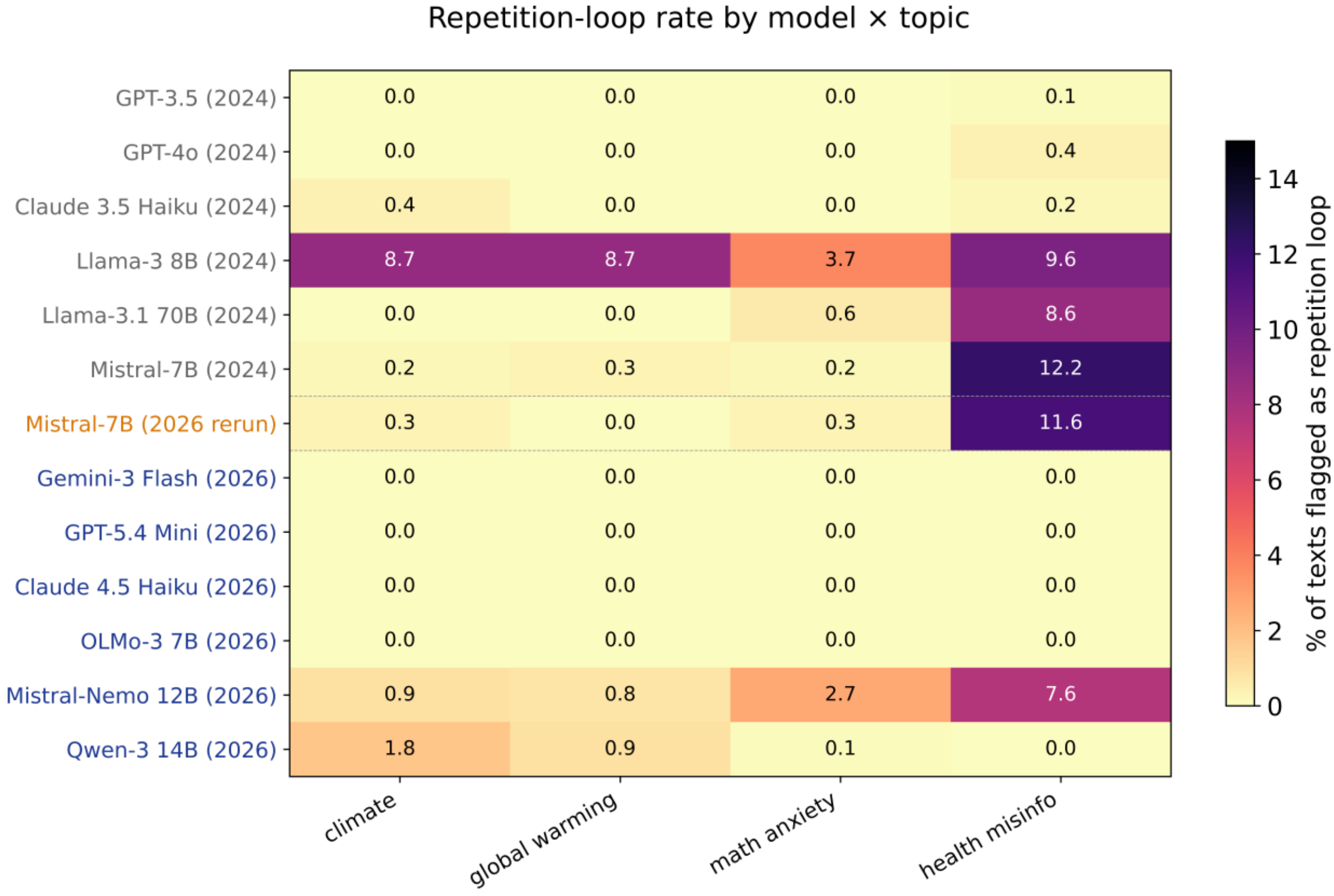

Fig. 4.1: The percentage of texts that were categorised as containing repetition loops, for the models and topics. Years refer to the year of when the datasets were generated (2024 vs 2026). Alt text: Shown is a heatmap of 13 models times 4 topics, showing the percentage of texts that were categorised as containing repetition loops. Most cells are near 0%; a few rows have values of up to 12.2%, mainly in the health-misinformation column.

(i) With respect to answer length and average sentence length, GPT-5.4-Mini reaches the 2,048-token generation cap in about 16% of health-misinformation texts. The general tendency is for earlier models to produce shorter answers overall but longer average sentences. For details, see Figure 4.2.

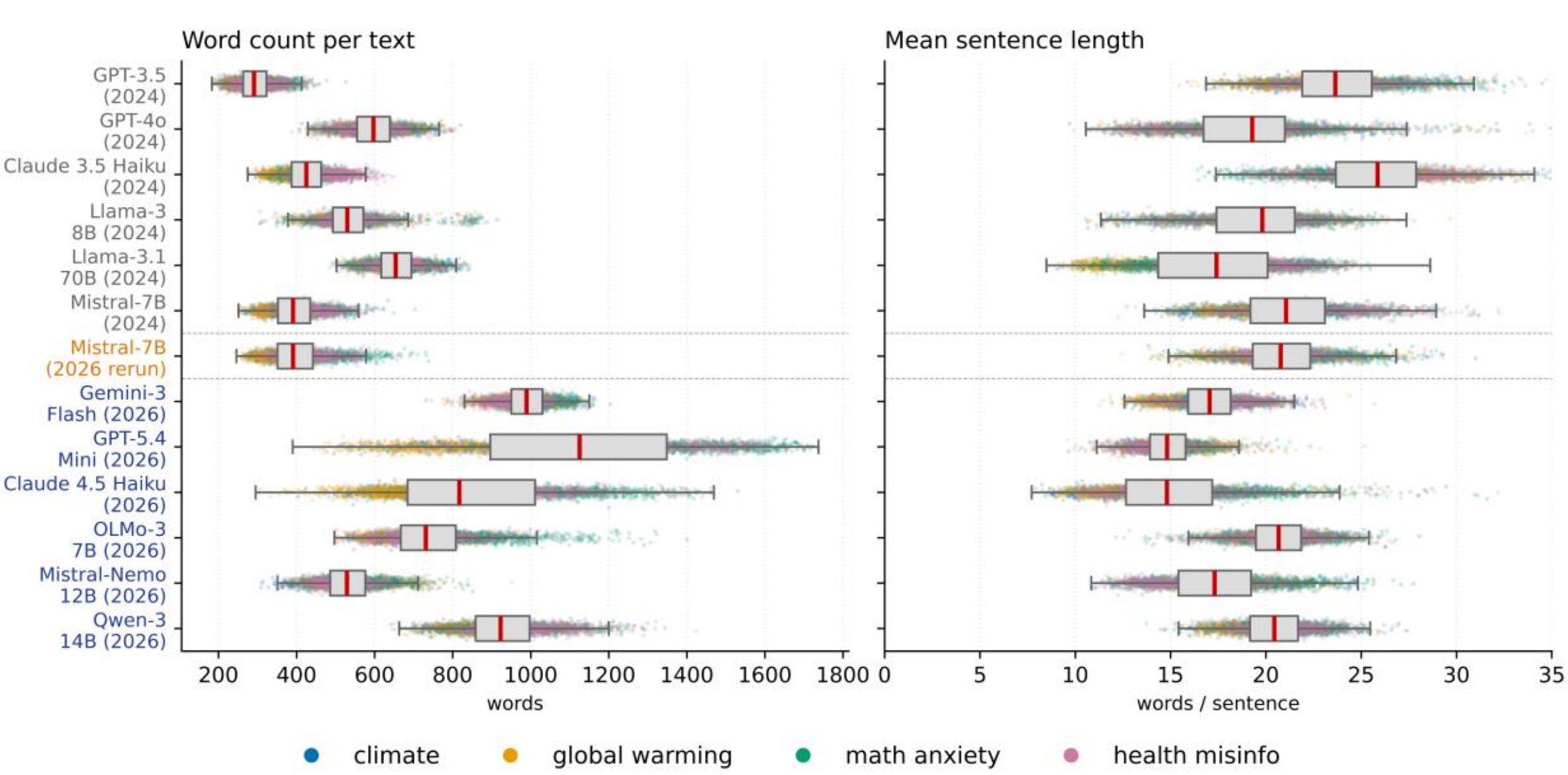

Fig. 4.2: Text length measured in words per text, and mean words per sentence, per model, with years of dataset generation.

Alt text: Shown are two box-plot panels, containing one row per model, with points coloured by topic, within a box’s whiskers. Word counts rise sharply for the 2026 models; average sentence length varies to a lesser degree.

(ii) With respect to lexical diversity, type-token ratio and vocabulary size are both confounded by length: TTR falls and vocabulary size grows as answers become longer. The general tendency is therefore for terse earlier models to score highest on TTR (GPT-3.5: approximately 0.5-0.6) and lowest on vocabulary size, with the verbose more current models showing the reverse. These rankings largely reflect output length rather than vocabulary richness. For details, see Figure 4.3.

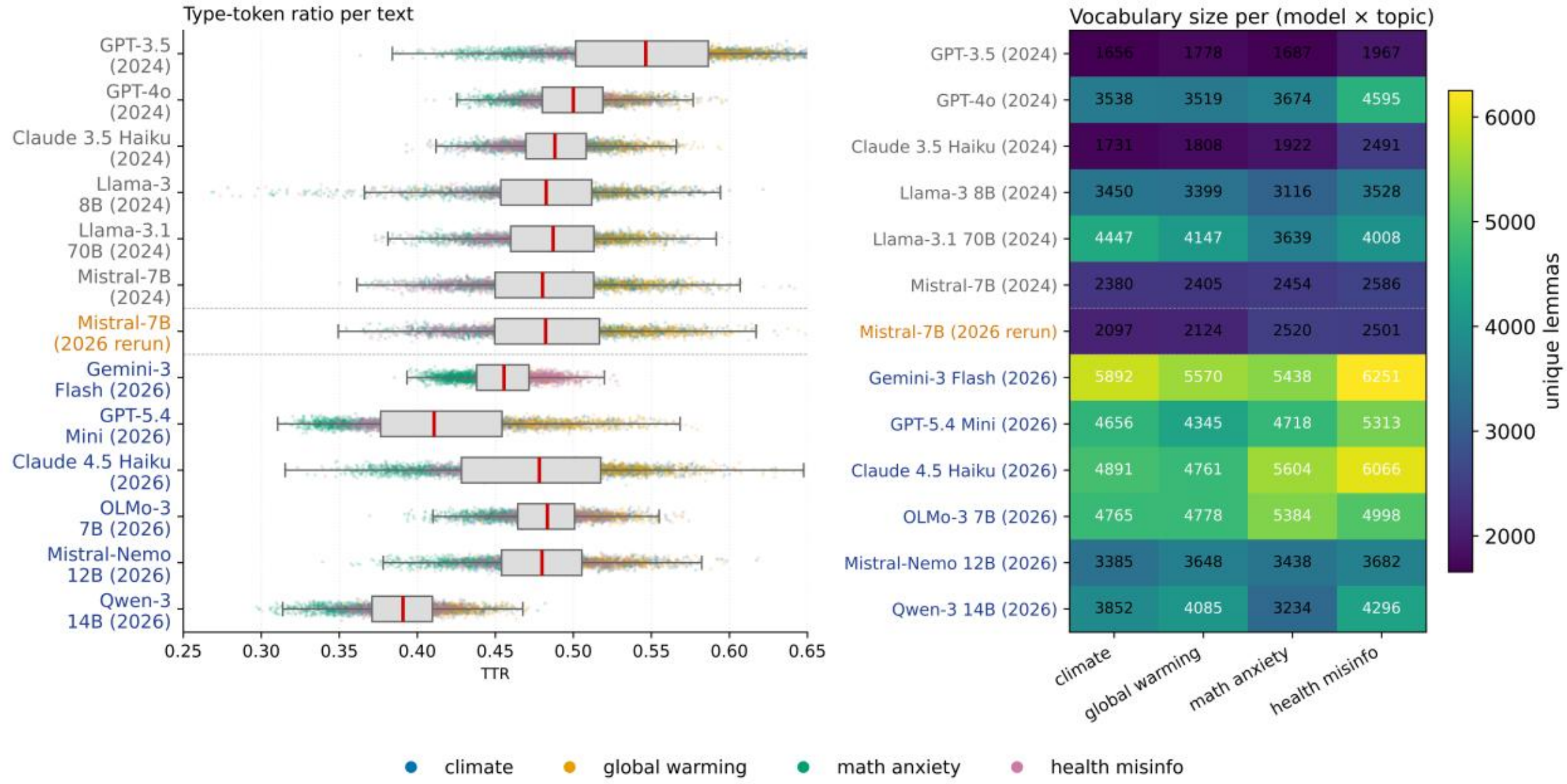

Fig. 4.3: Left: Type-token ratio (TTR) per text, per model. Right: Vocabulary size in unique lemmas per model and topic. Years in both graphs refer to year of data generation.

Alt text: Left: Shown are box plots of type-token ratio per model; whilst there is only some degree of variation, two models have a notably lower TTR, namely GTP-5.4 and Qwen-3, both of which are part of the 2026 cohort. Right: Shown is a heatmap for vocabulary-size, for the 13 models and the 4 topics. Brighter colours indicate larger vocabularies; the models in the 2026 cohort have notably larger vocabulary sizes than the models in the 2024 cohort.

(iii) Disclaimers behave differently. Rather than fading, safety disclaimers intensify in the Claude lineage (highest in Claude-4.5 Haiku: 46%) whilst sharply dropping in the GPT lineage (GPT-5.4 Mini: 0.2%). The open-weight current models (e.g. OLMo-3-7B, with 16% self-reference) are closer to the 2024 cohort than to their closed-API contemporaries. For details, see Figure 4.4. Refusal rates are virtually zero for all models and thus not visualised.

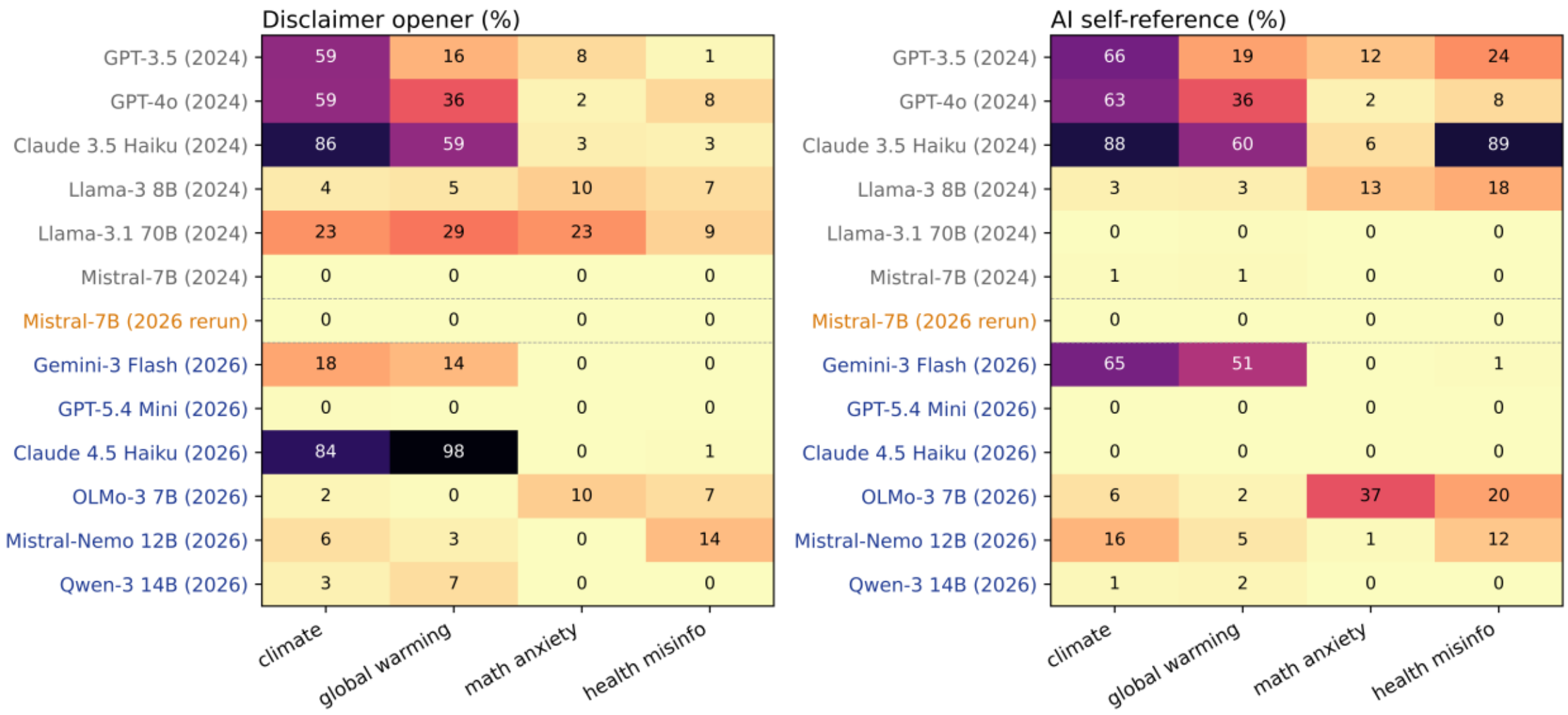

Fig. 4.4: Left: The percentage of texts opening with a safety disclaimer, per model and topic. Right: The percentage of texts containing an AI self-reference, per model and topic. Years in both graphs refer to year of data generation.

Alt text: Shown are two heatmaps with the percentage of texts opening with a safety disclaimer and the percentage of texts containing an AI self-reference, for the 13 models and the 4 topics. Higher rates are model-specific and concentrate on the climate topic and the global warming topic.

(iv) An analysis of non-prompt-matched human reference data (Reddit r/ChangeMyView opinion posts and a Reuters within-author dataset) is available on our GitHub repository (github.com/fsu-nlp/ai-idiolects). In general, within-speaker variation in a given domain is greater than within-model variation, while between-speaker variation is greater than between-model variation, which points in the same direction as research by Zamaraeva et al. (2025).

The findings described above show that individual models display both differences and similarities across the dimensions we examined. Some of these appear to be dependent on the specific cohort, such as output length, while others seem more dependent on the model itself, such as the presence of AI self-references. However, an analysis of each model's output points to the distinct characteristics present, even in features that result naturally from a writer's or model's habitual way of expression, rather than being consciously selected, such as average sentence length and lexical richness. At the same time, each model appears to be internally consistent in the way it expresses itself, as illustrated by the relatively tight distributions shown in Fig. 4.2, even though the models differ visibly from one another in their mean output length.

## 5. Stylometric principal component analysis

To examine whether the LLMs investigated in this study produce texts with recognisably distinct stylistic profiles – and whether those profiles cluster by model family, generation cohort, or neither – a principal component analysis (PCA) of word frequency data was conducted. To ensure that any stylistic differences reflect model behaviour rather than topic effects, both the 2024 cohort outputs (Improta et al., 2024) and the 2026 cohort outputs generated for the present study were drawn from a single topic domain: *health misinformation*. The analysis places these texts side by side within a single stylometric space, treating each model's characteristic word distribution as an analogue to an idiolect – the individual's distinctive and consistent way of using language.

### 5.1 Methodology

Stylometric analysis was conducted using the stylo package (v. 0.7.5) for R (Eder, Rybicki & Kestemont 2015), which implements principal component analysis (PCA) on word frequency profiles. The corpus comprised the subset of texts on the topic of health misinformation generated by twelve distinct large language models: six from the 2024 cohort (GPT-3.5, GPT-4o, Haiku, Llama-3-8B, Llama-3.1-70B, and Mistral-7b, sourced from Improta et al., 2024) and six from the 2026 cohort (Haiku-4-5, Gemini-3-flash, GPT-5-4-mini, Nemo-12b, Olmo3-7b, and Qwen3-14b, generated for the present study) to be accessed in the repository [fsu-nlp/ai-idiolects](fsu-nlp/ai-idiolects). Two additional files constituted a methodological replication control: independent reruns of the Mistral-7B-Instruct-v0.2 model executed in LM Studio with sampling parameters faithful to the ones used by Improta et al. (2024) but distinct random seeds, following the anchoring protocol described in Section 3.1. Prior to analysis, all corpus files were cleaned of CSV structural artefacts introduced during data export, including row-start prefixes (topic and model name fields), language markers (ENG), Llama instruction tokens (e.g., [/INST], <<SYS>>, <|eot_id|>), and markdown formatting symbols, using a purpose-written R cleaning script. This step ensured that the stylometric results reflected genuine linguistic output rather than metadata formatting.

The PCA was run on the 150 most frequent words (MFW), culled at 50% - meaning only words present in at least half of all corpus files were retained. This culling threshold was chosen to ensure analytical focus on words shared broadly across models, removing low-frequency or model-idiosyncratic tokens that could introduce noise into the stylistic comparison. Since all texts share the same topic, high-frequency topic vocabulary is present across all files; culling primarily eliminates rare or unevenly distributed tokens whose presence reflects individual prompt variation rather than consistent stylistic habit.

The analysis used a correlation matrix, which normalises for differences in overall word frequency between models. Each model was represented by the maximum number of non-overlapping 10,000-word samples available from its corpus. This sampling procedure yields multiple data points per model rather than one – ranging from 30 samples for the shortest corpus to 125 for the longest – enabling the 95% confidence ellipses displayed in Figure 5.1[4] to be interpreted as estimates of within-model stylistic consistency: a compact ellipse indicates a stable, self-consistent voice; a large or elongated ellipse indicates greater variability across texts generated by a given model. The variation in sample size across models is itself a consequence of differences in total output volume and does not affect the validity of the ellipse comparisons, as all models are represented by a sufficient number of points for robust ellipse estimation.

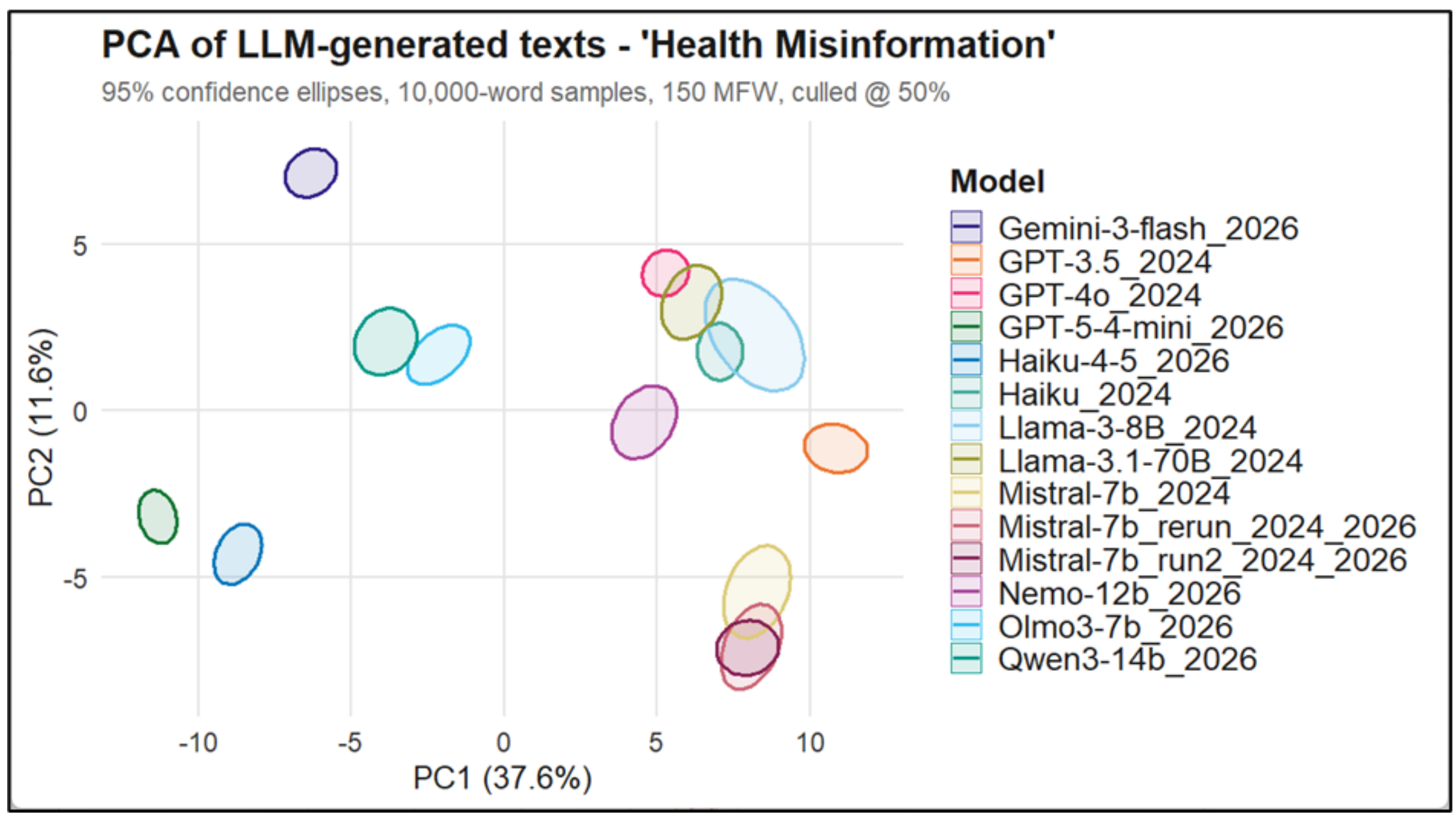

Fig. 5.1: PCA of LLM-generated texts on the topic of health misinformation.

Alt text: A two-dimensional principal component analysis (PCA) plot illustrating the stylistic distribution of text samples across fourteen distinct LLMs and two replication runs. The horizontal axis represents principal component 1 (PC1) and the vertical axis represents principal component 2 (PC2). Individual data points on the plot correspond to distinct 10,000-word text samples extracted from the health misinformation corpus. Data points are colour-coded by model family using a discrete, colour-blind safe palette, with each model group enclosed by a 95% confidence ellipse. The ellipses vary significantly in size, shape, and position: some models' outputs form highly compact, isolated ellipses indicating narrow within-model stylistic consistency, while other cohorts display larger, more elongated, or overlapping ellipses that indicate higher variation.

[4] Figure 5.1 was created using the ggplot2 package for R (Wickham 2016). Colours used are drawn from Paul Tol's colour-blind safe discrete rainbow palette (Tol 2021), selected to ensure accessibility for readers with colour vision deficiencies.

**5.2 Results**

The PCA results indicated that the first principal component (PC1) accounted for 37.6% of the total variance, while the second principal component (PC2) contributed a further 11.6%. Together, these two components captured just under half of the stylistic variation present in the corpus. The sharp drop from PC1 (37.6%) to PC2 (11.6%) suggests that PC1 represents a dominant, interpretable signal rather than distributed noise.

*Mistral replication: methodological validation*

The two independent reruns of Mistral-7B-Instruct-v0.2 (Run 5: mean PC1 = +8.05, mean PC2 = −7.07, n = 44; Run 6: mean PC1 = +7.93, mean PC2 = −7.12, n = 45), described in 3.1, landed in near-identical positions in the PCA space. This highly accurate replication suggests that the stylometric method is sensitive enough to recover consistent model signatures across independent generation runs with different random seeds.

Their proximity to the original Improta et al. Mistral-7b_2024 file (mean PC1 = +8.27, mean PC2 = −5.41) is also noticeable. The two reruns sit close to the original on PC1 with distances of approximately 1.67 and 1.74, respectively, driven primarily by a moderate shift on PC2 (−5.41 vs −7.07/−7.12). The minor PC2 differences could be a result of a combination of random variation and minor variation due to LM studio versioning that we could not control.

*PC1: The generational divide*

PC1 constitutes the primary axis of differentiation and aligns closely with generation year. Five of the six models of the 2026 cohort returned negative PC1 scores (range: −11.34 to −2.14), while all six 2024 models returned positive scores (range: +5.28 to +10.79). The exception was Nemo-12b_2026 (PC1 = +4.59), which positioned itself on the 2024 side of the axis. This separation was not imposed by the algorithm – the PCA had no knowledge of generation year – but emerged purely from word frequency patterns.

As the orientation of PCA axes is mathematically arbitrary and may differ depending on corpus composition, interpretations are based on the relative positions of models and the contrast between word loadings, that is, which words and models cluster together and which stand apart, rather than on whether individual loadings are positive or negative.

Words with the strongest loadings associated with the 2026 style include: *because, truth, real, is, they, most, are, do, uncertainty, when, not, but*, *than*. This vocabulary points to a direct and argumentative register: causal connectives (e.g., *because*), finite verbs (*is, are, do*), epistemic vocabulary (*truth, real, uncertainty*), and contrastive markers (*but, not*, *than*). The models of the 2026 data generation cohort appear to address the reader directly. The token *t*, a tokenisation artefact arising from the processing of contractions (e.g., *don't*, *can't*), also loads toward the 2026 style, which might be consistent with greater use of contracted forms in the more conversational register characteristic of 2026 cohort output. However, in order to obtain a more nuanced picture, we look at it in more detail in Section 6.

Among the highest-loading words associated with the 2024 style are: *and*, *theories*, *information*, *lead*, *to*, *individuals*, *sources*, *informed*, *conspiracy*, *spread*, *making*, *professionals*, *can*, *significant*, *promote*, *promoting, misinformation*, *treatments*. This vocabulary may be seen as reflecting a formal, hedged, nominalised academic register: abstract nouns (*information, theories, conspiracy*) and the connectives (*and, to, lead*). The 2024 cohort tends to write about health misinformation by means of accumulating noun phrases and cautious formulations rather than making direct argumentative statements.

Quantitatively, GPT-3.5_2024 occupied the most extreme 2024-style position (mean PC1 = +10.79, n = 30 samples), while GPT-5-4-mini_2026 occupied the most extreme 2026-style position (mean PC1 = −11.34, n = 125 samples). The contrast described above is most pronounced at the extremes of the PC1 loading spectrum; approximately one third of the 150 MFW show weak loadings close to zero and do not meaningfully distinguish between generation cohorts.

*PC2: Possible framing differences*

PC2 (11.6%) accounts for a substantially smaller share of variance than PC1 and should be interpreted with greater caution. The loading pattern suggests a possible secondary dimension relating to how health misinformation is rhetorically framed, though the pattern is less clear-cut than PC1 and rests partly on function words.

Among the words loading most strongly in the positive direction are: *must*, *the*, *an*, *of*, *like*, *a*, *science*, *scientific*, *society*, *where*, *literacy*, *content*, *public*, *social*. Several of these – notably *must*, *science*, *scientific*, *society*, and *literacy* – may be seen as pointing toward an institutional, society-level perspective concerned with public understanding and scientific communication. However, the four second-strongest loading words (*the*, *an*, *of*, *a*) are basic articles and prepositions that do not clearly signal style, and *like* – the fifth strongest – is ambiguous between comparative and informal usage.

On the other hand, words loading most strongly in the negative direction are: *false*, *serious*, *all*, *I*, *claims*, *about*, *harm*, *make*, *be*, *people*, *believe*, *think*, *attention*, *vaccines*. Lexical items such as *false*, *serious*, *claims*, *harm*, *believe*, *think*, and *I* may be seen as reflecting a more direct, engaged framing – naming specific harms, using first person, and addressing beliefs explicitly. Interestingly, *evidence* (−0.1201) also loads negatively, which is counterintuitive if the negative pole is characterised as confrontational; this may reflect its use in direct appeals rather than formal argumentation. As with the positive direction, several high-loading words (*about*, *be*, *all*, *some*) are basic function words without clear rhetoric implications.

Approximately half of the MFW wordlist shows weak PC2 loadings close to zero, including *misinformation* itself (−0.0119), and does not meaningfully distinguish between the two poles. Gemini-3-flash_2026 scored highest on this axis (mean PC2 = +7.12), followed by GPT-4o_2024 (+4.14) and Llama-3.1-70B_2024 (+3.29), while the Mistral family scored most negatively – Mistral-7b_2024 (−5.41) and both reruns (−7.07 and −7.12). Importantly, this dimension does not align with generation cohort: both the datasets belonging to the 2024 and 2026 cohorts appear on both sides of the axis, suggesting that if the framing difference is to be taken seriously, it reflects something else than the temporal shift captured by PC1.

## 6. Contractions – a nuanced investigation

To complement the broad and more general computational measures reported in Section 4 and the stylometric analysis in Section 5, a targeted corpus-linguistic investigation was conducted using a purpose-written R search script applied to the same cleaned corpus files used for the PCA (texts on health misinformation). The present section focuses on contracted forms as a case study. Contractions were selected for two reasons. First, they constitute a marker of informal[5], conversational register and are seen as "usually not appropriate in formal writing" according to Cambridge English Dictionary Online[6]. Second, the stylometric analysis identified a tokenisation artefact (*t*), see Section 5.2, loading toward the 2026 side of PC1, a pattern consistent with, but not conclusive of, greater contraction use in 2026 model output. The corpus-based analysis below allows this preliminary observation to be examined directly in the original texts, and, as shown below, to be substantially refined.

[5] Quirk and Greenbaum (1973) talk about "informal contractions" when describing phrases in question.

### 6.1 Methodology

Contracted forms were grouped into six categories according to the negator, verb or auxiliary being contracted: *'t* (negation: *don't, doesn't, can't, won't, isn't, aren't, didn't, wouldn't, couldn't, shouldn't, wasn't, weren't, haven't, hasn't, hadn't*), *'s* (*is/has*: *it's, that's, there's, here's, what's, who's*), *'m* (*am*: *I'm*), *'re* (*are*: *we're, you're, they're*), *'ve* (*have*: *I've, we've, you've, they've*), and *'d* (*would/had*: *I'd, we'd, you'd, they'd, he'd, she'd*). The purpose-written code only searched for contractions listed above, there was no interference with the genitive *'s* category. Both capitalised and non-capitalised forms were searched for.

Two categories are subject to a structural ambiguity that cannot be accounted for through string search alone. The *'s* category comprises *is* (*it's raining*) and *has* (*it's been raining*); the *'d* category comprises *would* (*I'd go*) and *had* (*I'd gone*). Disambiguating these would require dependency parsing rather than surface-form search, and the present results for these two categories should be read with this limitation in mind.

Raw counts and normalised frequencies per million words were calculated for each model using R, with normalisation applied to account for the substantial differences in corpus size across models' output files (ranging from approximately 300,000 to 1.26 million words).

### 6.2 Methodological challenges: the orthographic anomalies of LLM apostrophes

A significant technical challenge was discovered during manual checks of preliminary results in AntConc (Anthony 2022). While human writers typically default to standard straight (') or curly typographic apostrophes ('), the models in the present corpus used a range of alternative Unicode symbols, including:

- The standard straight apostrophe (', \u0027)
- The right single quotation mark / curly apostrophe (’, \u2019)
- The left single quotation mark (‘, \u2018)

[6] Cambridge English Dictionary. s.v. *contractions*. Access 29.06.2026. URL: https://dictionary.cambridge.org/grammar/british-grammar/contractions.

- The modifier letter apostrophe (ʼ, \u02bc)
- The acute accent (´, \u00b4)
- The grave accent (`, \u0060)
- The prime symbol (′, \u2032)
- The single high-reversed-9 quotation mark (‛, \u201b)

The fact that standard regular expression engines treat transitions between alphanumeric characters and specific Unicode punctuation marks as boundary markers caused search patterns to fail at capturing all instances. To ensure statistical integrity across the corpus, we implemented a custom regular expression pattern builder. This addition featured an expanded, multi-variant character class containing all eight discovered apostrophe glyphs. This methodology ensured complete, non-duplicate extraction of all contraction types selected for the present investigation.

### 6.3 Results

*Overall contraction frequency*

The data reveals a striking degree of stylistic variation across the evaluated models. At the upper extreme, Claude-Haiku-4-5 displays a frequency of over 30,000 contractions per million words, making it a distinct outlier among the LLMs tested. Conversely, models such as GPT-3.5 demonstrate an almost total avoidance of these forms, registering just 120.2 per million words. While there is an intuitive upward trend toward higher contraction frequencies in the newer 2026 cohort, the picture is very nuanced. The aggregate distributions across the health misinformation corpus are summarised in Table 6.1.

**Table 6.1**: Total contraction frequency per model (35 forms across 6 categories; health misinformation corpus).

| Model | Generation cohort | Raw count | Total words | Per million words |
|---|---|---|---|---|
| **Claude-Haiku-4-5** | 2026 | 28,997 | 947,247 | 30,611.9 |
| **Mistral-Nemo-Instruct** | 2026 | 3,939 | 524,826 | 7,505.3 |
| **Llama-3.1-70B** | 2024 | 3,585 | 651,595 | 5,501.9 |
| **OLMo-3-7B-Instruct** | 2026 | 3,513 | 715,838 | 4,907.5 |
| **Llama-3.8B** | 2024 | 2,396 | 528,248 | 4,535.8 |

| **Mistral-7b (rerun 6)** | Replication | 1,616 | 438,169 | 3,688.1 |
|---|---|---|---|---|
| **Mistral-7b (rerun 5)** | Replication | 1,577 | 434,358 | 3,630.6 |
| **Mistral-7b-Instruct** | 2024 | 1,285 | 434,716 | 2,955.9 |
| **Claude Haiku** | 2024 | 2,222 | 871,278 | 2,550.3 |
| **Qwen3-14B-Instruct** | 2026 | 2,186 | 967,433 | 2,259.6 |
| **Gemini-3-Flash** | 2026 | 1,503 | 961,258 | 1,563.6 |
| **GPT-5-4-Mini** | 2026 | 1,557 | 1,259,841 | 1,235.9 |
| **GPT-4o** | 2024 | 641 | 590,715 | 1,085.1 |
| **GPT-3.5** | 2024 | 36 | 299,484 | 120.2 |

*Category-level findings*

An individual look at each category reveals a much more precise picture than the combined figures suggest, and one that relates more directly to the PCA finding that motivated this investigation. The distributions and proportional shares are presented in Figure 6.1.

The *'t* (negation) category alone shows a marked generational divide: the 2026 cohort mean is 3,139.2 per million compared to 96.7 per million for 2024, a ratio of approximately 32:1. This pattern appears to be driven heavily by Claude-Haiku-4-5. However, if we exclude it, the remaining five models of the 2026 cohort still average 857.3 per million, nearly nine times the 2024 mean (96.7 per million words). Negation contractions, in other words, are markedly more frequent in 2026 model output.

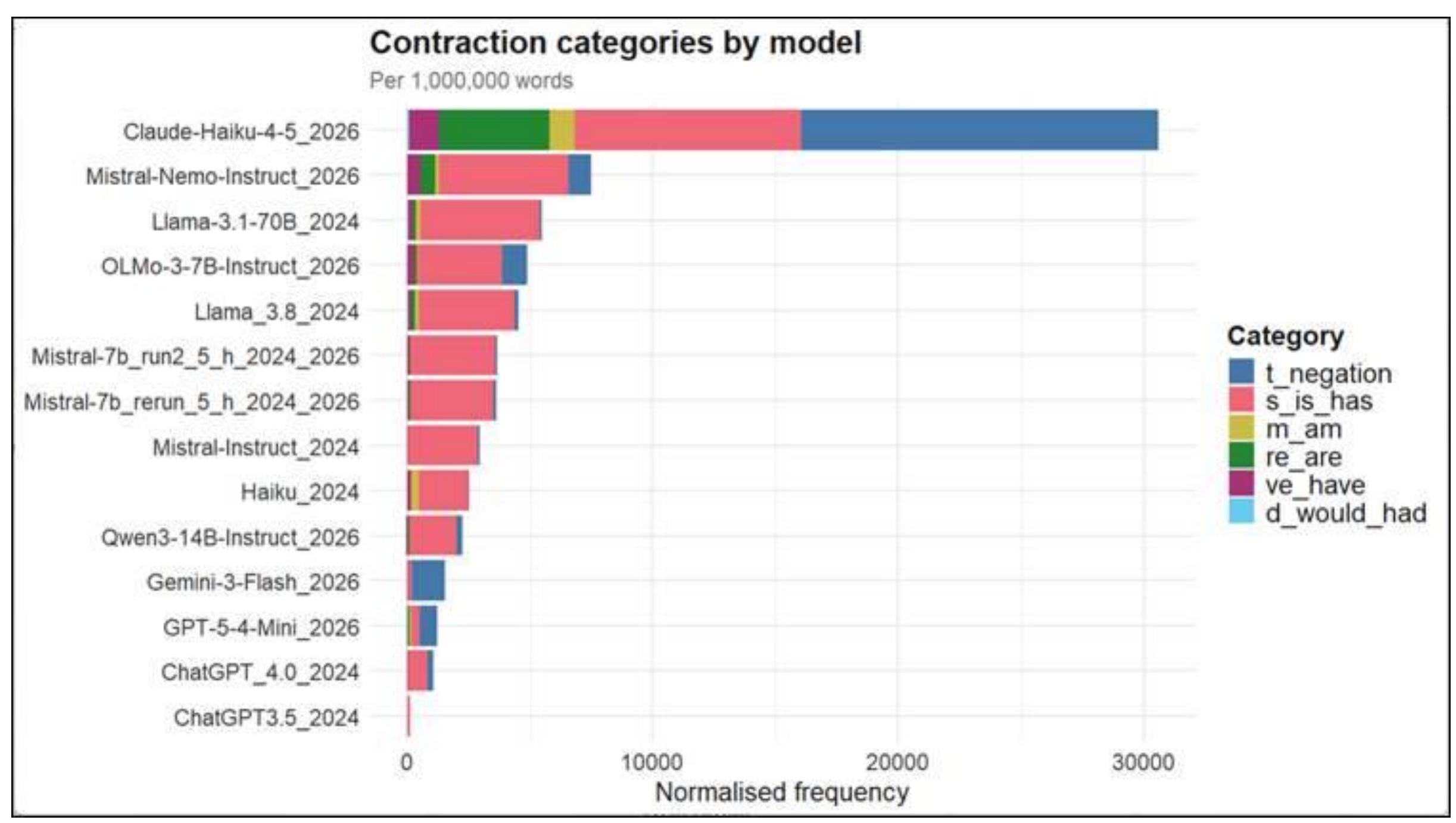

Fig. 6.1: Normalised frequencies across models for the present corpus of texts on health misinformation.
Alt text: A horizontal stacked bar chart titled "Contraction categories by model," displaying the normalised frequency of contraction usage per one million words across fourteen distinct large language models and two replication runs. The x-axis shows normalised frequency ranging from 120 to over 30,000 per million words.

The *'s* category shows a less pronounced pattern. While the 2026 mean (3,391.5 per million) is higher than the 2024 mean (2,419.9 per million), they differ by a ratio of 1.40:1 – a modest shift compared to the divergence seen in negation – and this category in fact accounts for the largest share of contracted forms in most individual models, regardless of cohort. The primary generational signal identified via PC1, then, remains strongly associated with negation contractions rather than with contractions as a general category.

Two further observations concern the consistency of individual models' contraction profiles. Gemini-3-Flash shows a markedly skewed profile: 1,357.6 per million for *'t* against 152.9 per million for *'s* – the inverse of the pattern observed in almost every other model, where *'s* dominates. Gemini's contraction use is therefore not significantly lower or higher than other models', but qualitatively different in kind, concentrated primarily in negation. Claude-Haiku-4-5's, by contrast, is elevated across both major categories simultaneously (14,548.5 per million for *'t*; 9,224.6 per million for *'s*), distinguishing it from models such as Mistral-Nemo-Instruct, which appears to contain many instances of *'s* (5,251.3 per million) while remaining comparatively moderate in *'t* (931.7 per million). The exact breakdown across all six examined contraction categories is shown below in Table 6.2.

**Table 6.2**: Contraction frequency by category and model (per million words).

| **Model** | **Generation year** | **'t (neg.)** | **'s (is/has)** | **'m** | **'re** | **'ve** | **'d** |
|---|---|---|---|---|---|---|---|
| **Claude-Haiku-4-5** | 2026 | 14,548.5 | 9,224.6 | 1,050.4 | 4,493.0 | 1,202.4 | 92.9 |
| **Mistral-Nemo-Instruct** | 2026 | 931.7 | 5,251.3 | 167.7 | 590.7 | 554.5 | 9.5 |
| **Llama-3.1-70B** | 2024 | 79.8 | 4,875.7 | 201.1 | 116.6 | 176.5 | 52.2 |
| **OLMo-3-7B-Instruct** | 2026 | 1,047.7 | 3,408.6 | 43.3 | 138.3 | 269.6 | 0.0 |

| **Llama-3.8B** | 2024 | 159.0 | 3,877.0 | 162.8 | 181.7 | 94.7 | 60.6 |
|---|---|---|---|---|---|---|---|
| **Mistral-7b (rerun 6)** | Replication | 93.6 | 3,471.3 | 0.0 | 47.9 | 75.3 | 0.0 |
| **Mistral-7b (rerun 5)** | Replication | 96.7 | 3,421.1 | 2.3 | 48.4 | 59.9 | 2.3 |
| **Mistral-7b-Instruct** | 2024 | 57.5 | 2,838.6 | 2.3 | 25.3 | 32.2 | 0.0 |
| **Claude Haiku** | 2024 | 28.7 | 2,040.7 | 308.7 | 4.6 | 167.6 | 0.0 |
| **Qwen3-14B-Instruct** | 2026 | 226.4 | 1,960.9 | 2.1 | 44.5 | 25.8 | 0.0 |
| **Gemini-3-Flash** | 2026 | 1,357.6 | 152.9 | 9.4 | 12.5 | 31.2 | 0.0 |
| **GPT-5-4-Mini** | 2026 | 723.1 | 350.8 | 62.7 | 77.8 | 7.9 | 13.5 |
| **GPT-4o** | 2024 | 242.1 | 783.8 | 1.7 | 25.4 | 32.2 | 0.0 |
| **GPT-3.5** | 2024 | 13.4 | 103.5 | 0.0 | 0.0 | 0.0 | 3.3 |

Individual models exhibit stable and distinctive contraction profiles that differ not only in overall frequency but in internal composition – which specific contraction types are favoured and in what proportion. The contrast between Claude-Haiku-4-5 (a markedly higher usage of contracted forms across categories), Gemini-3-Flash (a profile skewed toward negation), and GPT-5-4-Mini (maintaining a comparatively restrained profile) illustrates three qualitatively distinct contraction patterns and preferences within the same 2026 generation cohort. What the data shows, taken as a whole, is consistent with the central argument of the present chapter, and strengthens it beyond the picture based on the aggregate counts alone.

**7. Discussion and conclusions**

As justified above, the results described in the present work show that the notion of idiolect(s) can be applied to the output of LLM-based tools and constitutes a useful analytical framework for describing and analysing this output. To this end, the present section provides answers to the three research questions listed in Section 3, discusses the limitations of the approach, and identifies potential reasons for the stylistic differences observed between models documented in the analyses presented.

With respect to RQ1 – *in what sense is the notion of idiolect applicable?* The detected differences align well with the definition offered in Richards, Platt & Platt (1992: 172, see Section 2.1): "In its widest sense, someone's idiolect includes their way of communicating; for example, their choice of utterances and the way they interpret the utterances made by others.

In a narrower sense, an idiolect may include those features, either in speech or writing, which distinguish one individual from others." The evidence reviewed here suggests that LLM output meets both the broad and narrow criteria of this definition: models differ from one another in consistent ways across multiple linguistic levels, and do so without being instructed to write distinctively.

With respect to RQ2 – *on which linguistic levels do individual LLM-based tools exhibit stable and consistent differences?* The present chapter documents such differences across several levels of description, moving from the global to the local. At the broadest level, models differ systematically in output volume: word count per text and average sentence length (see Section 4) both show variation and similarities between models. At the stylistic level, contraction patterns (see Section 6), lexical richness (see Section 4) and the distribution of high-frequency function words, as shown by the PCA (see Section 5), reveal stable profiles that allow individual models to be distinguished from one another with a high degree of reliability. Also, the presence of model- and topic-specific openers (disclaimers and AI self-references) illustrates distinct "ways of speaking" (see Section 4).

With respect to RQ3, *do AI-idiolects change over time, and can we identify family-level continuities across successive generations?* The answer is complex. The dominant finding from the PCA (Section 7) done on health misinformation texts is that generation cohort – 2024 versus 2026 – is a stronger predictor of stylistic features than model family. The shift between generations appears to be consistent and measurable: 2024 models favour more formal and nominalised register, while 2026 models tend toward more direct and argumentative prose. This observed generational shift cannot be explained in a simple way, and a full account lies outside the scope of the present chapter. One plausible contributing factor concerns the sourcing of training data. If major AI developers draw on texts from the same data providers, shifts in the composition and characteristics of the datasets between training cycles could produce correlated stylistic changes across otherwise independent model families. This could potentially explain why the generational effect in the present study happens across families – affecting GPT, Claude, Gemini and others simultaneously – rather than affecting a single developer's models. A parallel from the human world could be cross-generational shifts in people's idiolects that happen, e.g., in migration contexts (Verhaeghe, Van Avermaet & Derluyn 2019) or which, according to, e.g., Marzona, we're facing now – with Generation Z preferring brief forms of communication and code-switching frequently (Marzona 2025). Another scenario for such a shift could be one in which one generation receives a substantially more thorough education than the preceding one. In such a scenario, a spectrum of idiolects among language users of that generation could also change collectively.

We believe that several factors contribute to the consistent linguistic profiles in LLM-based tools' outputs. Each model is trained on a unique dataset or specific combinations of datasets, shaping its vocabulary and syntactic and stylistic preferences. Reinforcement learning from human feedback (RLHF) further influences the model (Christiano et al. 2017; Stiennon et al. 2020; Ouyang et al. 2022), as raters' preferences leave traces in the model's output. Additionally, developers use system-level pre-prompts (Improta et al. 2024) to guide model behaviour toward specific styles which may be seen as a baseline constraint, but the resulting linguistic fingerprint forms the unintentional, non-negotiable idiolect that cannot be prompt-engineered away in a default LLM version. Together, these elements suggest that linguistic distinctiveness in LLMs is structurally embedded in how they are developed. On the other hand, it may also reflect emergent abilities (Wei et al. 2022, Bushwick, Musser, and Feder 2023) – skills not explicitly trained for which arise from the interplay of scale, data, and fine-tuning rather than from deliberate design decisions.

Naturally, it is worth acknowledging the limitations of this framing. The concept of idiolect has traditionally been applied to individual human speakers, whose linguistic patterns emerge from life experience gathered over time. LLM training is not life experience in that sense, and applying the term risks anthropomorphising LLM-based systems. Furthermore, a given model's linguistic profile remains largely fixed within a version, unlike the dynamic, continuously evolving nature of a human idiolect. Nevertheless, the core criteria that make idiolect a useful analytical concept – stability, individuality, uniqueness, and emergence rather than intentional selection – are precisely the properties that the evidence reviewed here suggests LLM output possesses. We would like to emphasise that the term idiolect is adopted in the present chapter not as a metaphysical claim about AI agency or personhood, but as an analytical framework for a phenomenon that existing terminology does not yet adequately capture.

## References

Amin, Madeeha, Hina Rasheed, Said Ali & Ahmun Ara. 2025. Artificial intelligence and the standardisation of Global English: A sociolinguistic inquiry. *Qualitative Research Journal for Social Studies* 2 (2). 365–375. https://doi.org/10.63878/qrjs310.

André, Christopher M. J., Helene F. L. Eriksen, Emil J. Jakobsen, Luca C. B. Mingolla & Nicolai B. Thomsen. 2023. Detecting AI authorship: Analyzing descriptive features for AI detection. In *NL4AI 2023: Seventh Workshop on Natural Language for Artificial Intelligence*, Rome, Italy. https://ceur-ws.org/Vol-3551/paper3.pdf.

Anthony, Laurence. 2022. *AntConc* (Version 4.2.0) [Windows]. Tokyo: Waseda University. https://www.laurenceanthony.net/software.html.

Baker, Paul. 2023. A year to remember? Introducing the BE21 corpus and exploring recent part of speech tag change in British English. *International Journal of Corpus Linguistics* 28. 407–429. https://doi.org/10.1075/ijcl.22007.bak.

Bergs, Alexander. 2025. Construction Grammar and Literature. In Miriam Fried & Kiki Nikiforidou (eds.), *The Cambridge handbook of construction grammar* [Cambridge Handbooks in Language and Linguistics], 623–647. Cambridge: Cambridge University Press. https://doi.org/10.1017/9781009049139.

Bitton, Yehonatan, Elad Bitton & Shai Nisan. 2025. Detecting stylistic fingerprints of large language models. *ArXiv preprint* https://doi.org/10.48550/arXiv.2503.01659

Bushwick, Sophie, George Musser & Elah Feder. 2023. Should we care about AI's emergent abilities? *Scientific American*. Should We Care About AI's Emergent Abilities? | Scientific American (accessed 28.06.2026).

Cambridge English Dictionary. n.d. s.v. contractions. https://dictionary.cambridge.org/grammar/british-grammar/contractions (accessed 29 June 2026).

Christiano, Paul. F., Jan Leike, Tom B. Brown, Miljan Martic, Shane Legg & Dario Amodei. 2017. Deep reinforcement learning from human preferences. *Advances in Neural Information Processing Systems* 30: 4302–4310. https://proceedings.neurips.cc/paper_files/paper/2017/hash/d5e2c0adad503c91f91df240d0cd4e49-Abstract.html.

Crothers, Evan N., Nathalie Japkowicz & Herna L. Viktor. 2023. Machine-generated text: A comprehensive survey of threat models and detection methods. *IEEE Access* 11: 70977–71002. https://doi.org/10.1109/ACCESS.2023.3294090.

Dai, Kun & Quanguo Liu. 2024. Leveraging artificial intelligence (AI) in English as a foreign language (EFL) classes: Challenges and opportunities in the spotlight. Computers in Human Behavior 159: 108354. 10.1016/j.chb.2024.108354.

Diessel, Holger. 2017. Usage-based linguistics. In Mark Aronoff (ed.), *Oxford Research Encyclopedia of Linguistics*. New York: Oxford University Press. https://doi.org/10.1093/acrefore/9780199384655.013.363.

Dunn, Jonathan. 2024. *Computational construction grammar: A usage-based approach*. Cambridge: Cambridge University Press. https://doi.org/10.1017/9781009233743.

Eder, Maciej, Jan Rybicki & Mike Kestemont. 2015. Stylometry with R: A package for computational text analysis. *The R Journal* 8 (1). 107–121. https://doi.org/10.32614/rj-2016-007.

Encyclopædia Britannica. n.d. Idiolect (In language: Definitions of language). *Britannica Online Encyclopedia*. https://www.britannica.com (accessed 7 February 2025).

Fried, Miriam & Kiki Nikiforidou (eds.). 2025. *The Cambridge handbook of construction grammar*. Cambridge: Cambridge University Press. https://doi.org/10.1017/9781009049139.

Geng, Mingmeng & Thierry Poibeau. 2025. On the detectability of LLM-generated text: What exactly is LLM-generated text? Ar*Xiv preprint:2510.20810v1*. https://doi.org/10.48550/arXiv.2510.20810.

Georgiou, Georgios P. 2025. Differentiating between human-written and AI-generated texts using automatically extracted linguistic features. *Information* 16 (11). 979. https://doi.org/10.3390/info16110979.

Herbold, Steffen, Annette Hautli-Janisz, Ute Heuer, Zlata Kikteva & Alexander Trautsch. 2023. A large-scale comparison of human-written versus ChatGPT-generated essays. *Scientific Reports* 13: 18617. https://doi.org/10.1038/s41598-023-45644-9.

Hickey, Raymond. 2012. Internally and externally motivated language change. In Juan Manuel Hernández-Campoy & Juan Camilo Conde-Silvestre (eds.), *The handbook of historical sociolinguistics*, 401–421. Malden: Wiley-Blackwell. https://doi.org/10.1002/9781118257227.ch21.

Hilpert, Martin. 2024. The road ahead for construction grammar. *Constructions and Frames* 16: 2, 255–277. https://doi.org/10.1075/cf.23014.hil

Honnibal, Matthew, Ines Montani, Sofie Van Landeghem, and Adriane Boyd. 2020. spacy: Industrial-strength natural language processing in python. Zenodo https://doi.org/10.5281/zenodo.1212303/

Hsiao, Jo-Chi, Shiowyun Chang & Jason S. Chang. 2025. AI for language learning: Opportunities and challenges. In Yu-Ju Lan, Grace Yue Qi & Dorothy Chun (eds.), *AI-mediated language education in the metaverse era* (Chinese Language Learning Sciences). Singapore: Springer. https://doi.org/10.1007/978-981-95-0245-5_8.

Improta, Riccardo, Giuseppe A. Veltri & Massimo Stella. 2024. TextualLLMap: A dataset of 28,000 Large Language Models' writings designed to expose their biases on societal issues. *OSF Preprint* https://doi.org/10.31234/osf.io/xwpe8.

Juzek, Thomas S. 2026. *LexA-Index: AI vs. human lexical overuse explorer*. https://www.aiwordexplorer.com/ (accessed 18 May 2026).

Kim, Jinhee. 2023. Leading teachers' perspective on teacher-AI collaboration in education. *Education and Information Technologies* 29 (7): 8693-8724. https://doi.org/10.1007/s10639-023-12109

Kobak, Dmitry, Rita González-Márquez, Emőke-Ágnes Horvát & Jan Lause. 2025. Delving into ChatGPT usage in academic writing through excess vocabulary. *Science Advances* 11. 10.1126/sciadv.adt3813.

Kumarage, Tharindu, Garima Agrawal, Paras Sheth, Raha Moraffah, Aman Chadha, Joshua Garland & Huan Liu. 2024. A survey of AI-generated text forensic systems: Detection, attribution, and characterization. *ArXiv preprint* https://doi.org/10.48550/arXiv.2403.01152

Kytö, Merja, John Scahill & Harumi Tanabe (eds.). 2010. *Language change and variation from Old English to Late Modern English: A festschrift for Minoji Akimoto*. Frankfurt: Peter Lang. https://doi.org/10.3726/978-3-0351-0092-1.

Litvinova, Tatiana. 2021. RusIdiolect: A new resource for authorship studies. In Tatiana Antipova (ed.), *Comprehensible Science: ICCS 2020*, 14–23. Cham: Springer International Publishing. https://doi.org/10.1007/978-3-030-66093-2_2.

Mair, Christian. 1998. Corpora and the study of major varieties in English: Issues and results. In H. Lindquist, S. Klintborg, M. Levin & M. Estling (eds.), *The major varieties of English: Papers from MAVEN 97*, 139–157. Växjö: Acta Wexionensia. https://searchworks.stanford.edu/view/4509309.

Mair, Christian. 2026. Uniformity and Diversity at the Same Time?: Artificial Intelligence and the Standardisation Paradox in Twenty-First-Century English. In Peter Collins & Adam Smith (eds.), *World-wide perspectives on English usage: Into the third millennium,* 228-246. Cambridge: Cambridge University Press. https://doi.org/10.1017/9781009569699.012.

Marzona, Yessy. 2025. Language Change in the Communication of Generation Z. (2025). Jurnal Ilmiah Langue and Parole, 8(2), 153-161. https://doi.org/10.36057/jilp.v8i2.734

Milička, Jiří, Anna Marklová & Václav Cvrček. 2025. Benchmark of stylistic variation in LLM-generated texts. *ArXiv preprint* https://arxiv.org/abs/2509.10179.

Millar, Robert M. 2024. *Linguistic contact and language change: An introduction*. Cambridge: Cambridge University Press. https://doi.org/10.1017/9781009071093

Muñoz-Ortiz, Alberto, Carlos Gómez-Rodríguez & David Vilares. 2024. Contrasting linguistic patterns in human and LLM-generated news text. *Artificial Intelligence Review* 57: 265. https://doi.org/10.1007/s10462-024-10903-2.

Naveed, Muhammad. 2024. *AI-generated text on diabetes: A dataset of ChatGPT and Gemini responses*. https://zenodo.org/records/14284509.

Ouyang, Long, Jeff Wu, Xu Jiang, Diogo Almeida, Carroll L. Wainwright, Pamela Mishkin, Chong Zhang, Sandhini Agarwal, Katarina Slama, Alex Ray, John Schulman, Jacob Hilton, Fraser Kelton, Luke Miller, Maddie Simens, Amanda Askell, Peter Welinder, Paul F. Christiano, Jan Leike & Ryan Lowe. 2022. Training language models to follow instructions with human feedback. *In Proceedings of the 36th International Conference on Neural Information Processing Systems (NIPS 22). Curran Associates Inc., Red Hook, NY, USA, Article 2011, 27730–27744.*35, 27730–27744. https://doi.org/10.48550/arXiv.2203.02155; https://proceedings.neurips.cc/paper/2022/hash/b1efde53be364a73914f58805a001731-Abstract-Conference.html.

Owan, Valentine Joseph, Kinsgley Bekom Abang, Delight Omoji Idika, Eugene Onor Etta, & Bassey Asuquo Bassey. 2023. Exploring the potential of artificial intelligence tools in educational measurement and assessment. *Eurasia Journal of Mathematics, Science and Technology Education* 19 (8): em2307. https://doi.org/10.29333/ejmste/13428.

Pemberton, Ian. 2024. Usage-based linguistics. In *Usage-based second language instruction*, 145–158. Cham: Palgrave Macmillan. https://doi.org/10.1007/978-3-031-53414-0_10.

Quirk, Randolph & Sidney Greenbaum. 1973. *A university grammar of English*. London: Longman. https://api.semanticscholar.org/CorpusID:60988901.

Rao, Zixin, Youssef Mohamed, Shang Liu & Zeyan Liu. 2025. Two birds with one stone: Multi-task detection and attribution of LLM-generated text. *ArXiv preprint* https://doi.org/10.48550/arXiv.2508.14190.

Reviriego, Pedro, Javier Conde, Elena Merino-Gómez, Gonzalo Martínez & José Alberto Hernández. 2024a. *Humans vs ChatGPT texts on TOEFL questions and HC3 dataset* [Data set]. Zenodo. https://doi.org/10.5281/zenodo.11199030 (accessed 2 January 2025).

Reviriego, Pedro, Javier Conde, Elena Merino-Gómez, Gonzalo Martínez & José Alberto Hernández. 2024b. Playing with words: Comparing the vocabulary and lexical diversity of ChatGPT and humans. *Machine Learning with Applications* 18. 100602. https://doi.org/10.1016/j.mlwa.2024.100602.

Richards, Jack C, John Platt, Heidi Platt. 1992. *Longman Dictionary of Language Teaching and Applied Linguistics*. Longman Group UK Limited.

Rodriguez Louro, Celeste. 2025. AI systems are built on English – but not the kind most of the world speaks. *The University of Western Australia*. https://www.uwa.edu.au/news/article/2025/may/ai-systems-are-built-on-english-but-not-the-kind-most-of-the-world-speaks.

Rudnicka, Karolina. 2023. Can Grammarly and ChatGPT accelerate language change? AI-powered technologies and their impact on the English language: Wordiness vs. conciseness. *Procesamiento del Lenguaje Natural* 71, 205–214. https://doi.org/10.26342/2023-71-16.

Rudnicka, Karolina. 2025a. Each AI chatbot has its own, distinctive writing style—just as humans do. *Scientific American*. https://www.scientificamerican.com/article/chatgpt-and-gemini-ai-have-uniquely-different-writing-styles/ (accessed 9 July 2025).

Rudnicka, K. (2025b, February). The language of ai tools as idiolects– thus comparable to other idiolects: https://doi.org/10.31219/osf.io/7qwzu_v1.

Rühlemann, Christoph. & Martin Hilpert. 2017. Colloquialization in journalistic writing: The case of inserts with a focus on well. *Journal of Historical Pragmatics* 18: 104–135. https://doi.org/10.1075/jhp.18.1.05ruh.

SensorTower. 2026. *2026 state of AI*. https://sensortower.com/blog/state-of-ai-2026 (accessed 22 June 2026).

Shijaku, Rexhep. 2022. ChatGPT-generated text detection corpus. GitHub. https://github.com/rexshijaku/chatgpt-generated-text-detection-corpus/tree/main (accessed 2 January 2025).

Singh, Satya Vir & Kamal Kant Hiran. 2022. The Impact of AI on Teaching and Learning in Higher Education Technology. *Journal of Higher Education Theory and* Practice 22 (13): 135–148. https://doi.org/10.33423/jhetp.v22i13.5514.

Smith, Genevieve, Eve Fleisig, Ishita Rustagi & Xavier Yin. 2025. Standard Language Ideology in AI-Generated Language. *ArXiv preprint* https://doi.org/10.48550/arXiv.2406.08726.

Stiennon, Nisan, Long Ouyang, Jeffrey Wu, Daniel M. Ziegler, Ryan Lowe, Chelsea Voss, Alec Radford, Dario Amodei & Paul F. Christiano. 2020. Learning to summarize from human feedback. *Advances in Neural Information Processing Systems* 33, 3008–3021. https://proceedings.neurips.cc/paper/2020/hash/1f89885d556929e98d3ef9b86448f951-Abstract.html.

Stokel-Walker, Chris. 2024. Chatbots have thoroughly infiltrated scientific publishing. *Scientific American*. URL https://www.scientificamerican.com/article/chatbots-have-thoroughly-infiltrated-scientific-publishing/ (accessed 01.07.2026).

Terčon, Luka & Kaja Dobrovoljc. 2025. Linguistic characteristics of AI-generated text: A survey. *ArXiv preprint:* https://doi.org/10.48550/arXiv.2510.05136.

Tol, Paul. 2021. Colour schemes. *SRON Technical Note SRON/EPS/TN/09-002*. https://sronpersonalpages.nl/~pault.

Verhaeghe, Floor, Piet Van Avermaet, and Ilse Derluyn. 2019. "Meanings Attached to Intergenerational Language Shift Processes in the Context of Migrant Families." Journal of

Ethnic and Migration Studies 48 (1): 308–26. https://doi.org/10.1080/1369183X.2019.1685377.

Wei, Jason, Yi Tay, Rishi Bommasani, Colin Raffel, Barret Zoph, Sebastian Borgeaud, Dani Yogatama, Maarten Bosma, Denny Zhou, Donald Metzler, Ed H. Chi, Tatsunori Hashimoto, Oriol Vinyals, Percy Liang, Jeff Dean & William Fedus. 2022. Emergent Abilities of Large Language Models. *Transactions on Machine Learning Research*. https://openreview.net/forum?id=yzkSU5zdwD.

Wickham, Hadley. 2016. *ggplot2: Elegant graphics for data analysis*. New York: Springer-Verlag. https://doi.org/10.1007/978-3-319-24277-4.

Wojan, Katarzyna. 2024. *Polska leksykografia autorska* [Polish authorial lexicography]. Gdańsk: Wydawnictwo Uniwersytetu Gdańskiego. https://wydawnictwo.ug.edu.pl/produkt/polska-leksykografia-autorska.

Wu, Junchao, Shu Yang, Runzhe Zhan, Yulin Yuan, Lidia Sam Chao & Derek Fai Wong. 2025. A Survey on LLM-Generated Text Detection: Necessity, Methods, and Future Directions. *Computational Linguistics* 51(1): 275–338. https://doi.org/10.1162/coli_a_00549

Zamaraeva, Olga, Dan Flickinger, Francis Bond, and Carlos Gómez-Rodríguez. 2025. Comparing LLM-generated and human-authored news text using formal syntactic theory. In Proceedings of the 63rd Annual Meeting of the Association for Computational Linguistics (Volume 1: Long Papers), 9041–9060, Vienna, Austria. Association for Computational Linguistics. https://doi.org/10.18653/v1/2025.acl-long.443

Zanotto, Sergio E. & Segun Aroyehun. 2024. Human variability vs. machine consistency: A linguistic analysis of texts generated by humans and large language models. *ArXiv preprint* https://doi.org/10.48550/arXiv.2412.03025.

Zhang, Mengxuan & Peter Crosthwaite. 2026. More human than human? Differences in lexis and collocation within academic essays produced by ChatGPT-3.5 and human L2 writers. *International Review of Applied Linguistics in Language Teaching* 64(2), 733-760 https://doi.org/10.1515/iral-2024-0196.